\pdfoutput=1

\documentclass[11pt]{article}

\usepackage{acl}

\usepackage{times}
\usepackage{latexsym}

\usepackage[T1]{fontenc}

\usepackage[utf8]{inputenc}

\usepackage{microtype}

\usepackage{graphicx}
\usepackage{bm}
\usepackage{amsmath}
\usepackage{amsfonts}

\title{Single Model Ensemble for Subword Regularized Models \\ in Low-Resource Machine Translation}

\author{Sho Takase \and Tatsuya Hiraoka \and Naoaki Okazaki \\
  Tokyo Institute of Technology \\
  \texttt{\{sho.takase@nlp., tatsuya.hiraoka@nlp., okazaki@\}c.titech.ac.jp} \\}

\begin{document}
\maketitle
\begin{abstract}
Subword regularizations use multiple subword segmentations during training to improve the robustness of neural machine translation models.
In previous subword regularizations, we use multiple segmentations in the training process but use only one segmentation in the inference.
In this study, we propose an inference strategy to address this discrepancy.
The proposed strategy approximates the marginalized likelihood by using multiple segmentations including the most plausible segmentation and several sampled segmentations.
Because the proposed strategy aggregates predictions from several segmentations, we can regard it as a single model ensemble that does not require any additional cost for training.
Experimental results show that the proposed strategy improves the performance of models trained with subword regularization in low-resource machine translation tasks.
\end{abstract}

\section{Introduction}
\label{sec:intro}
Subword regularizations are the technique to make a model robust to segmentation errors by using multiple subword segmentations instead of only the most plausible segmentation during the training process~\cite{kudo-2018-subword,provilkov-etal-2020-bpe}.
Previous studies demonstrated that subword regularizations improve the performance of LSTM-based encoder-decoders and Transformers in various machine translation datasets, especially in low-resource settings~\cite{kudo-2018-subword}.

However, previous subword regularizations contain the discrepancy between the training and inference.
In the training process, we stochastically re-segment a given sequence into subwords based on statistics such as the uni-gram language model~\cite{kudo-2018-subword}.
Thus, we use multiple segmentations for each input sequence.
In contrast, we use only the most plausible segmentation in the inference phase.
We expect that we can improve the performance by solving this discrepancy.

\begin{figure*}[!t]
  \centering 
  \includegraphics[width=16cm]{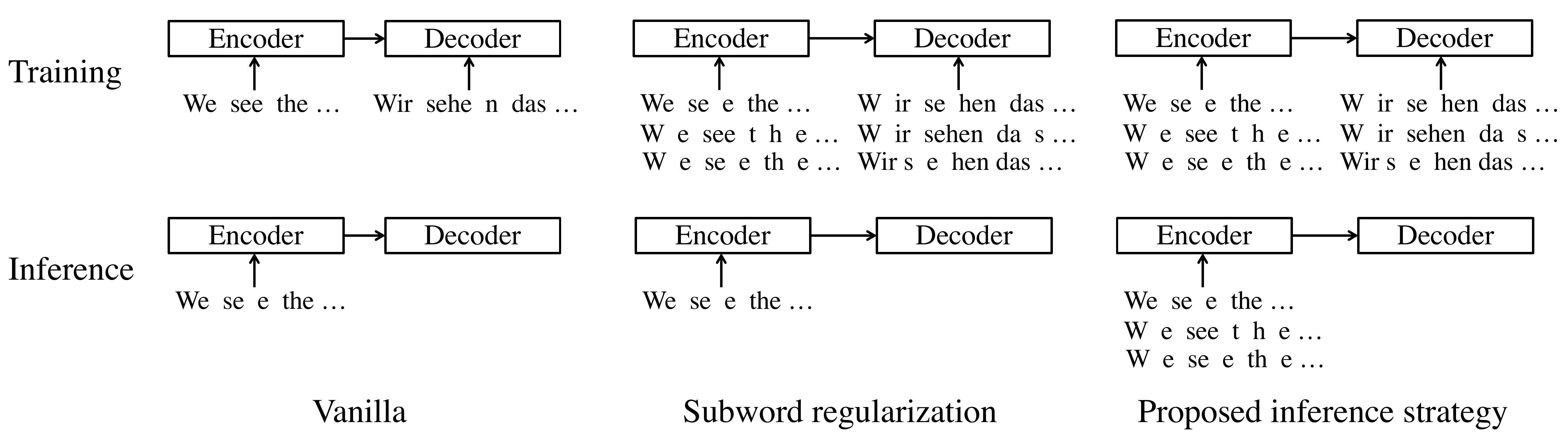}
   \caption{Overview of previous methods and the proposed inference strategy for an English-German pair ``We see the ...'' and ``Wir sehen das ...''. In the vanilla setting, we use the most plausible segmentation only in the training and inference. In the subword regularization, we use multiple segmentations during training but use only the most plausible segmentation in the inference phase. In the proposed inference strategy, we use multiple segmentations in both the training and inference phases.}
   \label{fig:overview}
\end{figure*}

To solve this discrepancy, we propose an inference strategy that uses multiple subword segmentations.
We construct multiple subword segmentations for an input in the same manner as that in the training process, and then aggregate the predictions from each segmentation.
Therefore, our proposed inference strategy can be regarded as a single model ensemble using multiple segmentations.
Figure \ref{fig:overview} illustrates the overview of previous methods and our proposed inference strategy.

We conduct experiments on several machine translation datasets.
Experimental results show that the proposed strategy improves the performance of a subword regularized model without any additional costs in the training procedure when the subword regularization significantly contributes to the performance, i.e., in low-resource settings.
Moreover, we indicate that our strategy can be combined with a widely used model ensemble technique.

\section{Subword Regularization}

Our proposed strategy is based on a model trained with subword regularization.
Thus, we briefly describe subword regularization in this section.

\newcite{kudo-2018-subword} proposed subword regularization to improve the robustness of a neural machine translation model.
Let $X$ and $Y$ be the source and target sentences, $\mathbf{x} = (x_1, ..., x_S)$ and $\mathbf{y} = (y_1, ..., y_T)$ be the most plausible subword segmentations corresponding to $X$ and $Y$.
In the vanilla training strategy, i.e., without subword regularization, we train the parameters of a neural machine translation model $\bm{\theta}$ to maximize the following log-likelihood:
\begin{align}
\mathcal{L}(\bm{\theta}) &= \sum_{(X, Y) \in \mathcal{D}} \log P(\mathbf{y} | \mathbf{x}; \bm{\theta}), \\
P(\mathbf{y}|\mathbf{x}; \bm{\theta}) &= \prod_{t=1}^{T} P(y_t | \mathbf{x}, \mathbf{y}_{<t}; \bm{\theta}),
\end{align}
where $\mathcal{D}$ is the training data and $\mathbf{y}_{<t} = (y_1, ..., y_{t-1})$.

In contrast, subword regularization uses multiple subword segmentations during training.
Let $P(\mathbf{x'} | X)$ and $P(\mathbf{y'} | Y)$ be segmentation probabilities for sequences $X$ and $Y$, respectively.
We optimize the parameters $\bm{\theta}$ with the following marginalized likelihood in subword regularization:
\begin{align}
\label{eq:subreg}
\mathcal{L'}(\bm{\theta}) &= \sum_{(X, Y) \in \mathcal{D}}
\mathbb{E}_{\substack{\mathbf{x'} \sim P(\mathbf{x'}|X) \\
\mathbf{y'} \sim P(\mathbf{y'} | Y)}}
[\log P(\mathbf{y'}|\mathbf{x'}; \bm{\theta})].
\end{align}
Because the number of possible segmentations increases exponentially with respect to the sequence length, it is impractical to optimize Equation (\ref{eq:subreg}) exactly.
Thus, \newcite{kudo-2018-subword} approximated Equation (\ref{eq:subreg}) with sampled segmentations from $P(\mathbf{x'} | X)$ and $P(\mathbf{y'} | Y)$,
\begin{align}
\label{eq:sampled_subreg}
\mathcal{L'}(\bm{\theta}) &\cong \sum_{(X, Y) \in \mathcal{D}} \log P(\mathbf{y}_j | \mathbf{x}_i; \bm{\theta}), \\
\mathbf{x}_i &\sim P(\mathbf{x'}|X), \\
\mathbf{y}_j &\sim P(\mathbf{y'}|Y).
\end{align}
We sample $\mathbf{x}_i$ and $\mathbf{y}_j$ for every mini-batch during training to yield a good approximation.

In the inference phase, we input the most plausible segmentation $\mathbf{x}$ and search a sequence $\mathbf{y}^*$ that maximizes the log-likelihood $\log P(\mathbf{y} | \mathbf{x}; \bm{\theta})$.
In other words, we input one segmentation to the model\footnote{\newcite{kudo-2018-subword} also proposed $n$-best decoding. This strategy uses $n$ segmentations but inputs them separately. In other words, a model receives only one segment and generates the corresponding output $n$ times in this strategy. We compare this strategy in experiments.} even though we use multiple segmentations during training.

\section{Proposed Method}
\subsection{Proposed Inference Strategy}
\begin{table}[!t]
  \centering
  \begin{tabular}{ c | c | c | c | c} \hline
  Language & Vocab & Train & Dev & Test \\ \hline
  En-De & 6K & 160K & 7283 & 6750 \\
  En-Vi & 4K & 133K & 1553 & 1268 \\ \hline
  \end{tabular}
  \caption{Details of each dataset.}
  \label{table:mt_dataset}
\end{table}

As described, previous subword regularizations use multiple segmentations during training but only one segmentation in the inference.
To solve this discrepancy, we propose an inference strategy that uses multiple segmentations as inputs.
In the proposed strategy, we search a sequence $\mathbf{y}^*$ that maximizes the following approximated marginalized likelihood:
\begin{align}
&\sum_{k=1}^{n} \log P(\mathbf{y} | \mathbf{x}_k; \bm{\theta}), \\
\mathbf{x}_k &= \begin{cases}
   \mathbf{x}  & k = 1 \\
   \mathbf{x}_i \sim P(\mathbf{x'} | X) &\textrm{Otherwise}.
   \end{cases}
\end{align}
In short, we approximate the marginalized likelihood in Equation (\ref{eq:subreg}) with the most plausible segmentation and sampled $n-1$ segmentations.

\begin{table*}[!t]
  \centering{}
  \begin{tabular}{ l | c | c | c | c } \hline
  Method & En-De & De-En & En-Vi & Vi-En \\ \hline \hline
  \multicolumn{5}{c}{Single Model} \\ \hline \hline
  Vanilla & 28.89 & 34.87 & 31.09 & 31.43 \\
  + w/ subword regularization (1) & 29.51 & 35.53 & 31.86 & \textbf{31.60} \\
  (1) + $n$-best decoding &  29.59 & 35.55 & 31.94 & 31.44 \\
  (1) + Proposed strategy & \textbf{29.72} & \textbf{35.68} & \textbf{32.16} & \textbf{31.60} \\ \hline \hline
  \multicolumn{5}{c}{Model Ensemble} \\ \hline \hline
  Vanilla & 30.03 & 36.04 & 32.22 & 32.46 \\
  + w/ subword regularization (2) & 30.83 & 36.83 & 33.22 & 32.83 \\
  (2) + $n$-best decoding & 30.81 & 36.83 & 33.29 & 32.76 \\
  (2) + Proposed strategy & \textbf{30.86} & \textbf{36.95} & \textbf{33.44} & \textbf{33.04} \\ \hline
  \end{tabular}
  \caption{BLEU scores on English-German and English-Vietnamese datasets.\label{tab:exp_main}}
\end{table*}

\subsection{Relation to Model Ensemble}
We often apply the model ensemble technique to achieve better performance~\cite{barrault-etal-2019-findings}.
In the model ensemble, we aggregate the predictions from $M$ models as follows:
\begin{align}
 \sum_{m=1}^{M} \log P(\mathbf{y} | \mathbf{x}; \bm{\theta}_m),
\end{align}
where $\bm{\theta}_m$ denotes parameters of the $m$-th model.

In comparison to this model ensemble, the proposed strategy does not use multiple models but aggregates predictions from multiple segmentations.
Thus, our proposed strategy can be regarded as the single model ensemble with multiple inputs.
In addition, we can combine the proposed strategy with the model ensemble.
We investigate the effect of this combination through experiments.

\section{Experiments}
\label{sec:exp}

\subsection{Datasets}
\newcite{kudo-2018-subword} reported that subword regularization is especially effective in low-resource settings.
Thus, we focus on low-resource machine translation tasks.
We used IWSLT 2014 English-German (En-De) data in the same pre-processing manner as \newcite{DBLP:journals/corr/RanzatoCAZ15}\footnote{\href{https://github.com/pytorch/fairseq/blob/master/examples/translation/}{github.com/pytorch/fairseq/blob/master/examples/translation/}} because this dataset is widely-used as the low-resource setting~\cite{sennrich-zhang-2019-revisiting,takase-kiyono-2021-rethinking}.
In addition, we used IWSLT 2015 English-Vietnamese (En-Vi) data which were pre-processed by \newcite{Luong-Manning:iwslt15}\footnote{\href{https://nlp.stanford.edu/projects/nmt/}{https://nlp.stanford.edu/projects/nmt/}}.

We used SentencePiece~\cite{kudo-richardson-2018-sentencepiece} to construct a vocabulary set.
We set the vocabulary sizes to 6k and 4k for En-De and En-Vi, respectively.
Table \ref{table:mt_dataset} summarizes the dataset sizes.

\subsection{Methods}
We used Transformer~\cite{NIPS2017_7181} as our encoder-decoder architecture because Transformers are widely used as strong baselines in sequence-to-sequence problems including machine translation.
We investigate the performance of the following configurations.

\noindent{\textbf{Vanilla}}: We trained Transformer~\cite{NIPS2017_7181} without subword regularization.
For hyper-parameters, we adopted the IWSLT setting in \texttt{fairseq}\footnote{\href{https://github.com/pytorch/fairseq}{https://github.com/pytorch/fairseq}}~\cite{ott-etal-2019-fairseq}.

\noindent{\textbf{Subword regularization}}: We trained Transformer, whose hyper-parameters are identical to Vanilla, with subword regularization.
We set the hyper-parameter $\alpha$ for sampling segmentations in subword regularization $0.2$ in the same as \newcite{kudo-2018-subword}.

\noindent{\textbf{$n$-best decoding}}: \newcite{kudo-2018-subword} proposed $n$-best decoding that generates $n$ sequences corresponding to $n$-best segmentations and then outputs the most plausible sequence.
We used this strategy for the model trained with subword regularization in the inference phase.

\noindent{\textbf{Proposed}}: We applied the proposed strategy to the model trained with subword regularization.
To ensure fair comparison, we used the identical number, $n = 5$, for the number of sampled segmentations and $n$-best decoding.

\subsection{Results}
Table \ref{tab:exp_main} shows BLEU scores of each configuration.
For each configuration, we trained three models with different random seeds, and reported the averaged scores except for the proposed strategy.
When we used the proposed strategy, we generated sequences three times with different random seeds for each model\footnote{Because the generated sequence mainly depends on the trained model, our inference strategy generates almost the same sequences even if we vary random seeds for samplings. However, we reported the averaged BLEU of $9$ sequences to make the results more reliable.}, and averaged the $9$ ($3$ models $\times$ $3$ sequences) scores.
Table \ref{tab:exp_main} also indicates BLEU scores with the ensemble of the above $3$ models.

For the single model setting, Table \ref{tab:exp_main} shows that subword regularization improved BLEU scores in all language pairs.
In particular, subword regularization gained more than 0.5 BLEU score from Vanilla except for Vi-En.
In these language pairs, the proposed strategy provided further improvements.
The proposed strategy achieved better performance than $n$-best decoding when we used the same number of segmentations as inputs.
Thus, our proposed method is more effective as the inference strategy.
Moreover, our strategy maintained the score in Vi-En although $n$-best decoding degraded the score slightly.
Therefore, the proposed strategy had no negative effect on the inference.

For the model ensemble setting, Table \ref{tab:exp_main} indicates that subword regularization also improved BLEU scores in all language pairs.
In this setting, the proposed strategy also provided further improvements in all language pairs.
Thus, the proposed strategy is effective even if we conduct the model ensemble technique.

\section{Performance in Enough Training Data}
\label{sec:large_data}
\begin{figure}[!t]
  \centering 
  \includegraphics[width=8cm]{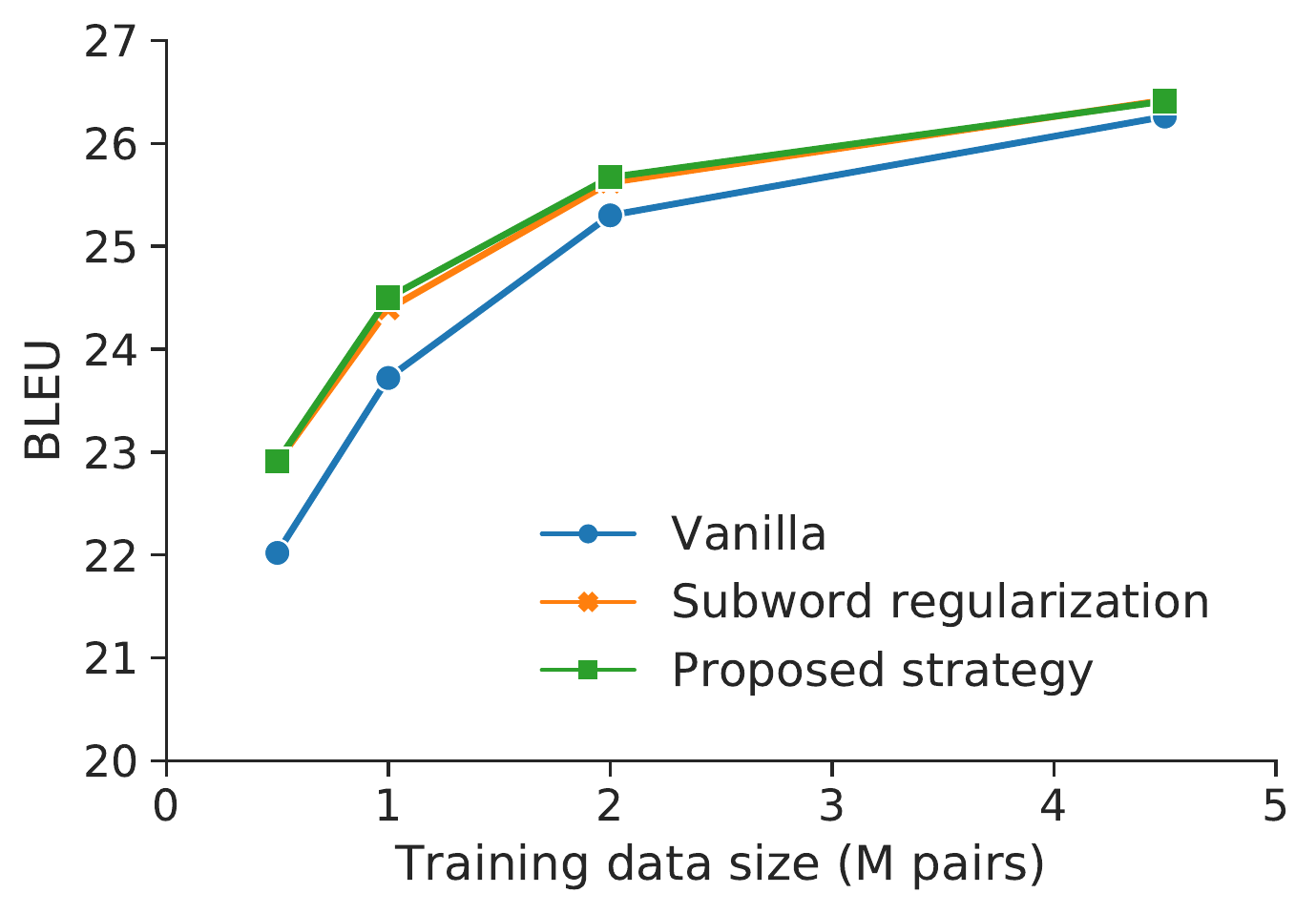}
   \caption{BLEU scores on newstest2013 when we vary the training data size.}
   \label{fig:score_datasize}
\end{figure}
Section \ref{sec:exp} shows the results in low-resource settings but previous studies reported that subword regularizations can improve the performance if we use sufficient training data.
Thus, we investigate the performance of subword regularization and proposed strategy by varying the size of training data.

We used the WMT 2016 English-to-German training dataset, which is widely used in previous studies~\cite{NIPS2017_7181,provilkov-etal-2020-bpe,ott-etal-2018-scaling}.
This dataset contains 4.5M sentence pairs, that are more than 25 times as many as IWSLT datasets.
We conducted pre-processing in the same manner as that in \newcite{ott-etal-2018-scaling}.
We trained the Transformer (base) model in \newcite{NIPS2017_7181}.
For subword regularization, we set $\alpha = 0.5$ in the same as \newcite{kudo-2018-subword}.
We evaluated BLEU scores on newstest2013, which is widely used as a valid data.

Figure~\ref{fig:score_datasize} shows BLEU scores of each method for each training data size.
This figure indicates that the model trained with subword regularization outperformed Vanilla in all training data sizes but the improvement decreased in accordance with the increase in the training data.
The proposed strategy slightly improved the performance from subword regularization for the small training data but the improvement also decreased as the training data increased.
When we used the entire training data (4.5M translation pairs), the BLEU score of the proposed strategy was identical to that of subword regularization.
This result implies that the impact of the proposed strategy on the performance is small when the improvement by subword regularization is small.
In other words, the proposed strategy is effective especially in low-resource settings because subword regularization probably provides much improvement in low-resource settings.
However, we emphasize that the proposed strategy has no negative effect on the BLEU score for sufficient training data fortunately.

\section{Related Work}
In this study, we proposed the inference strategy to mitigate the discrepancy between the training and inference in subword regularizations.
In experiments, we focused the subword regularization proposed by \newcite{kudo-2018-subword} but we can apply the proposed inference strategy to variants of the subword regularization such as BPE dropout~\cite{provilkov-etal-2020-bpe} and compositional word replacement~\cite{hiraoka-etal-2022}.
\newcite{takase-kiyono-2021-rethinking} reported that simple perturbations such as word dropout are effective in a large amount of training data.
Thus, we might improve the performance of the model trained with such simple perturbations if we use multiple inputs constructed by the same perturbation during the inference.

We focused on an input of a neural encoder-decoder.
In contrast, \newcite{NIPS2016_076a0c97} focused on internal layers.
For neural network methods, we often apply the dropout during the training but do not use it in the inference.
\newcite{NIPS2016_076a0c97} proposed the variational inference to mitigate this gap on the dropout.

As described in Section \ref{sec:intro}, our proposed inference strategy can be regarded as a single model ensemble.
\newcite{DBLP:journals/corr/HuangLPLHW17} and \newcite{kuwabara-etal-2020-single} also proposed single model ensemble methods.
\newcite{DBLP:journals/corr/HuangLPLHW17} proposed the snapshot ensemble that uses multiple models in the middle of the training.
\newcite{kuwabara-etal-2020-single} used pseudo-tags and predefined distinct vectors to obtain multiple models virtually during the training of a single model.
Since these methods are orthogonal to ours, we can combine our proposed strategy.

\section{Conclusion}
We proposed an inference strategy to address the discrepancy between the training and inference in subword regularizations.
Our proposed strategy uses multiple subword segmentations as inputs to approximate the marginalized likelihood used as the objective function during training.
The proposed strategy improved the performance of the model trained with subword regularization in cases where subword regularization provided the significant improvement, i.e., in low-resource settings.
Moreover, the proposed strategy outperformed the $n$-best decoding strategy~\cite{kudo-2018-subword}.
Experimental results show that our proposed strategy has no negative effect on the BLEU score even if the improvement by subword regularization is small.
Because the proposed inference strategy does not require any additional training cost, we encourage using the strategy to highlight the potential of models trained with subword regularization.

\section*{Ethical Considerations}
\noindent{\textbf{Limitations}}:
The proposed method improves the performance of encoder-decoders in the inference phase in the situation where subword regularizations are effective.
Thus, if subword regularizations are ineffective, the proposed method also might be ineffective.
Since subword regularizations are especially effective when the training data size is small~\cite{hiraoka-etal-2021-joint}, the proposed method is effective in low-resource settings.
In contrast, as in Section \ref{sec:large_data}, the improvements of both methods are small when we have an enough training data.

\noindent{\textbf{Risks}}:
Since the proposed method uses the standard neural encoder-decoder architecture without any modification, the proposed method also contains the risks of neural encoder-decoders.
For example, the under translation, that ignores some information in a source sentence during the translation, might happen.

\section*{Acknowledgements}
This work was supported by JSPS KAKENHI Grant Number JP21K17800 and JST ACT-X Grant Number JPMJAX200I.
The first author is supported by Microsoft Research Asia (MSRA) Collaborative Research Program.

\bibliographystyle{acl_natbib}




\end{document}